\documentclass[conference]{IEEEtran}
\IEEEoverridecommandlockouts

\usepackage[T1]{fontenc}
\usepackage{lmodern}
\usepackage{times}

\usepackage{amsmath,graphicx,hyperref,amssymb}
\usepackage{url}            
\usepackage{booktabs}       
\usepackage{amsfonts}       
\usepackage{nicefrac}       
\usepackage{microtype}      
\usepackage{titletoc}
\usepackage{pdfpages}
\usepackage{setspace}
\usepackage{subcaption}
\usepackage{graphicx}
\usepackage{multirow}
\usepackage{color}
\usepackage{colortbl}
\usepackage{xcolor}
\usepackage{setspace}
\usepackage{makecell}
\usepackage{cleveref}
\usepackage{algorithm}
\usepackage{balance}
\usepackage{algorithmicx}
\usepackage{algpseudocode}
\usepackage{pifont}         
\usepackage{amssymb}        
\usepackage{etoc}
\usepackage{amsthm}
\usepackage{balance}
\usepackage{xr-hyper}
\definecolor{lightkeycolor}{RGB}{255,240,240}
\definecolor{lightgray}{RGB}{234,234,234}

\begin{document}

\title{Collaborative Adaptive Curriculum for Progressive Knowledge Distillation}

\author{
\IEEEauthorblockN{Jing Liu\textsuperscript{1,2,3,$\dagger$},
Zhenchao Ma\textsuperscript{2,$\dagger$},
Han Yu\textsuperscript{1,$\dagger$},
Bobo Ju\textsuperscript{5},
Wenliang Yang\textsuperscript{1},
Chengfang Li\textsuperscript{4,*},
Bo Hu\textsuperscript{1,*},
Liang Song\textsuperscript{1,*}}
\thanks{\textsuperscript{$\dagger$}Equal contribution. \textsuperscript{*}Corresponding authors. This work is supported in part by the National Key Research and Development Program of China under Project No. 2024YFE0200700 (Subject No. 2024YFE0200703).
This work was also supported in part by the Specific Research Fund of the Innovation Platform for Academicians of Hainan Province under Grant YSPTZX202314, in part by the Shanghai Key Research Laboratory of NSAI and the Joint Laboratory on Networked AI Edge Computing, Fudan University-Changan.
}
\IEEEauthorblockA{\textsuperscript{1}Fudan University,
\textsuperscript{2}The University of British Columbia,
\textsuperscript{3}Duke Kunshan University,\\
\textsuperscript{4}Suzhou Inst. of Biomed. Eng. and Tech., Chinese Academy of Sciences,
\textsuperscript{5}Shanghai Shentong Metro Co., Ltd.}
}

\maketitle
\begin{abstract}
Recent advances in collaborative knowledge distillation have demonstrated cutting-edge performance for resource-constrained distributed multimedia learning scenarios. However, achieving such competitiveness requires addressing a fundamental mismatch: high-dimensional teacher knowledge complexity versus heterogeneous client learning capacities, which currently prohibits deployment in edge-based visual analytics systems. Drawing inspiration from curriculum learning principles, we introduce Federated Adaptive Progressive Distillation (\texttt{FAPD}), a consensus-driven framework that orchestrates adaptive knowledge transfer. \texttt{FAPD} hierarchically decomposes teacher features via PCA-based structuring, extracting principal components ordered by variance contribution to establish a natural visual knowledge hierarchy. Clients progressively receive knowledge of increasing complexity through dimension-adaptive projection matrices. Meanwhile, the server monitors network-wide learning stability by tracking global accuracy fluctuations across a temporal consensus window, advancing curriculum dimensionality only when collective consensus emerges. Consequently, \texttt{FAPD} provably adapts knowledge transfer pace while achieving superior convergence over fixed-complexity approaches. Extensive experiments on three datasets validate \texttt{FAPD}'s effectiveness: it attains 3.64\% accuracy improvement over FedAvg on CIFAR-10, demonstrates 2$\times$ faster convergence, and maintains robust performance under extreme data heterogeneity ($\alpha$=0.1), outperforming baselines by over 4.5\%.
\end{abstract}

\begin{IEEEkeywords}
Federated Learning, knowledge distillation, curriculum learning, multimedia hierarchy, collaborative intelligence
\end{IEEEkeywords}

\section{Introduction}
\label{sec:intro}
Federated Learning (FL) enables privacy-preserving visual model training on decentralized multimedia data~\cite{pfeiffer2023federated}, critical for visual recognition applications on resource-constrained edge devices in domains like mobile image classification~\cite{liu2025networking}, healthcare~\cite{wang2024mmoral,li2024decoding}, and video surveillance~\cite{liu2022learning}. A core challenge in FL is managing statistical heterogeneity~\cite{ehsanhallaji2024decentralized} and communication overhead while accommodating limited client resources~\cite{liu2024vertical,liu2026edgecloud}. To address these issues, Collaborative Knowledge Distillation (CKD) allows smaller client models to learn from a powerful, centralized teacher model~\cite{li2019fedmd}, thereby transferring complex knowledge without sharing raw data while striking a balance between model performance and the practical constraints of decentralized systems.

Despite its potential, CKD often fails due to a fundamental mismatch between the complexity of teacher-provided knowledge and the learning capacity of heterogeneous clients~\cite{pfeiffer2023federated}. Many existing methods~\cite{dai2023tackling} attempt to transfer a full, high-dimensional teacher representation from the start of training, yet such ``one-size-fits-all'' approaches can overwhelm clients with limited resources, leading to unstable training and poor generalization~\cite{seo2024relaxed,wu2025clipae}. Other approaches use rigid, predefined curricula, but these static schedules cannot adapt to the dynamic learning states of clients or the collective network~\cite{wang2023fedcda,wu2025survey}, proving ineffective in diverse federated settings.

The critical gap in current CKD methods is the absence of an adaptive mechanism to orchestrate knowledge transfer based on the network's collective learning progress. Data-free distillation approaches~\cite{zhang2022fedftg} synthesize training samples but employ fixed knowledge complexity. Other methods~\cite{zhao2024data-free} distill aggregated statistics without progressive adaptation, while~\cite{chen2023best} aggregates hyper-knowledge at constant dimensionality. An intelligent system must pace the introduction of complex information, ensuring foundational knowledge is mastered before more intricate details are introduced~\cite{liu2025networking,liu2025survey}. Accomplishing this requires a mechanism to monitor the network-wide learning state (i.e., consensus) and use this signal to dynamically adjust the curriculum. Without such adaptability, federated systems cannot efficiently transfer knowledge or accommodate client heterogeneity, leaving performance potential untapped.

Existing federated learning can be broadly categorized based on its knowledge transfer strategies~\cite{yan2024clientsupervised,wen2024diffimpute}. Traditional approaches like FedAvg~\cite{mcmahan2017communicationefficient} focus on parameter aggregation without leveraging external teacher knowledge, often struggling with statistical heterogeneity. To address this, FKD methods~\cite{li2019fedmd} introduce teacher supervision but typically transfer full, high-dimensional representations throughout training, which can overwhelm resource-constrained clients. More recently, curriculum-based strategies~\cite{wang2022federated} have attempted to structure learning by scheduling data or client participation. However, these methods predominantly rely on linear, predetermined schedules that fail to adapt to the dynamic, collective learning states of the network~\cite{liu2025anomaly,liu2026enhancing}, leaving a critical gap in orchestrating knowledge complexity according to real-time system capacity.

Recent advancements have explored integrating large-scale generative models and curriculum strategies into federated learning. Works such as FedDifRC~\cite{wang2025feddifrc} and VQ-FedDiff~\cite{yoon2025vqfeddiff} demonstrate techniques for adapting text-to-image diffusion models to federated settings, addressing communication efficiency. Meanwhile, curriculum learning approaches have been applied to schedule data samples~\cite{bengio2009curriculum} or client participation~\cite{wang2022federated,kang2025cufl}. Other efforts focus on parameter-efficient adaptation using prompt tuning~\cite{bao2024prompt} or collaborative pre-training~\cite{seo2024relaxed}. While these studies highlight the potential of sophisticated teacher models and curriculum strategies, they do not resolve how to progressively distill complex high-dimensional feature representations to resource-constrained clients~\cite{liu2026projecting}.

To bridge this gap, we propose Federated Adaptive Progressive Distillation (\texttt{FAPD}), a novel framework that orchestrates collaborative knowledge transfer through a consensus-driven curriculum. Unlike static approaches that overwhelm clients with complex information or rigid schedules that ignore learning dynamics, \texttt{FAPD} dynamically adapts the complexity of distilled knowledge to the network's collective capacity. Our approach introduces a hierarchical decomposition strategy that structures teacher features into ordered levels of importance, prioritizing fundamental patterns before intricate details in a manner analogous to human educational curricula~\cite{liu2025multimodala}. A central controller monitors network-wide learning stability and advances the curriculum only when consensus readiness is achieved, thereby ensuring that clients in heterogeneous environments are not destabilized by premature complexity. By synchronizing knowledge transfer with collective progress, \texttt{FAPD} effectively balances the trade-off between learning stability and performance, offering a robust solution for real-world federated systems where client resources and data distributions vary significantly.
Our contributions are summarized as follows:
\begin{itemize}
    \item We propose \texttt{FAPD}, a consensus-driven curriculum framework that dynamically orchestrates knowledge complexity in collaborative distillation by monitoring network-wide stability signals and adapting transfer pace accordingly.
    \item We design a PCA-based hierarchical knowledge decomposition that structures teacher features into variance-ordered principal components, enabling progressive distillation synchronized with heterogeneous client learning.
    \item We conduct extensive experiments that demonstrate \texttt{FAPD} achieves substantial improvements over baselines in accuracy, convergence speed, and training stability under diverse heterogeneous federated learning environments.
\end{itemize}

\begin{figure*}[t]
    \centering
    \includegraphics[width=\textwidth]{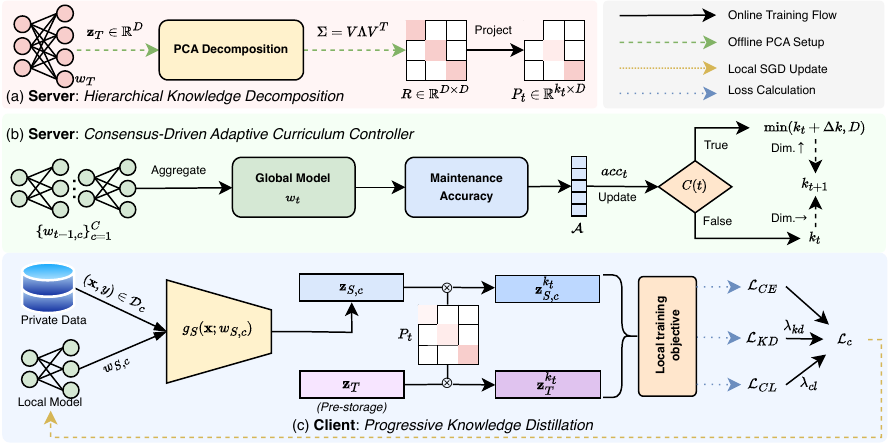}
    \caption{Illustration of \texttt{FAPD} featuring three components: (a) server-side hierarchical knowledge decomposition via PCA generating rotation matrix $R$, (b) consensus-driven adaptive curriculum controller monitoring accuracy history $\mathcal{A}$ to determine dimension $k_t$, and (c) client-side progressive knowledge distillation projecting features through $P_t$ to compute multi-objective loss $\mathcal{L}_c$.}
    \label{fig:framework_overview}
    \vspace{-15px}
\end{figure*}

\section{Related Work}
\label{sec:related}
\noindent\textbf{Collaborative Knowledge Distillation}. CKD enables resource-constrained clients to learn from server models while preserving data privacy. Early methods required public datasets~\cite{li2019fedmd}, while data-free alternatives synthesize samples via generative models~\cite{zhang2022fedftg} or distill aggregated statistics~\cite{chen2023best}. Recent frameworks address architectural heterogeneity through group distillation~\cite{dai2023tackling} or robust transfer protocols~\cite{wang2024mmoral}. However, existing approaches uniformly employ fixed-complexity transfer strategies, ignoring dynamic client learning capacity. In contrast, \texttt{FAPD} introduces an adaptive curriculum over distilled knowledge complexity.

\noindent\textbf{Curriculum Learning in Federated Settings}. Curriculum Learning improves convergence by training on progressively difficult examples~\cite{bengio2009curriculum}. Within FL, prevailing methods schedule data samples by difficulty metrics~\cite{wang2024mgr3}, manage client participation~\cite{wang2022federated}, or guide personalization from generic to client-specific features~\cite{kang2025cufl}. Nevertheless, such methods schedule data or clients rather than knowledge complexity. \texttt{FAPD} introduces curriculum over distilled knowledge dimensionality, paced by network-wide consensus signals reflecting collective learning states.

\section{Preliminary}
\label{sec:background}
\noindent\textbf{Federated Knowledge Distillation}. In a FL system with $C$ clients, each client $c \in \{1, \dots, C\}$ holds a private dataset $\mathcal{D}_c = \{(\mathbf{x}_i, y_i)\}_{i=1}^{n_c}$. The global objective is to minimize a weighted average of local loss without centralizing data:
\begin{equation}
    \min_{w} F(w) := \sum_{c=1}^{C} p_c F_c(w),
 \label{eq:fl_objective}
\end{equation}
where $p_c = n_c / \sum_j n_j$ is the weight for client $c$, and $F_c(w) = \frac{1}{n_c} \sum_{(\mathbf{x}, y) \in \mathcal{D}_c} \mathcal{L}(f(\mathbf{x}; w), y)$ is the local empirical risk. In FKD, a powerful teacher model with parameters $w_T$ assists the training of smaller client (student) models with parameters $w_{S,c}$. Knowledge is transferred by aligning feature representations. Let $g_T(\cdot; w_T)$ and $g_S(\cdot; w_{S,c})$ be the feature extractors of the teacher and student models, respectively. The local objective for each client $c$ is a combination of a standard classification loss and a distillation loss:
\begin{equation}
    \min_{w_{S,c}} \mathbb{E}_{(\mathbf{x}, y) \sim \mathcal{D}_c} \left[ \mathcal{L}_{CE}(f_S(\mathbf{x}; w_{S,c}), y) + \lambda \mathcal{L}_{KD} \right],
 \label{eq:fkd_objective}
\end{equation}
where $f_S$ is the student's prediction function, $\mathcal{L}_{CE}$ is the cross-entropy loss, and $\mathcal{L}_{KD} = \text{Dist}(g_S(\mathbf{x}; w_{S,c}), g_T(\mathbf{x}; w_T))$ is a loss function, such as Mean Squared Error, that aligns student and teacher features.

\noindent\textbf{Hierarchical Knowledge Representation via PCA}. A primary challenge in FKD is that high-dimensional teacher representations, $\mathbf{z}_T = g_T(\mathbf{x}; w_T) \in \mathbb{R}^D$, can overwhelm student models. To address this, knowledge can be structured and introduced progressively. PCA provides a principled method for decomposing a feature space into a hierarchy of components ordered by their contribution to the data's variance~\cite{li2024decodinga,li2024silent}. Given a set of teacher features, PCA finds an orthogonal transformation to a new coordinate system of principal components, where projecting a feature vector $\mathbf{z}_T$ onto the first $k$ principal components provides a compressed representation that captures the most significant variations in the data. By projecting $\mathbf{z}_T$ onto the subspace spanned by the eigenvectors $\{\mathbf{v}_1, \dots, \mathbf{v}_k\}$ of the data's covariance matrix (corresponding to the $k$ largest eigenvalues), we obtain the $k$-dimensional representation:
\begin{equation}
    \mathbf{z}_T^{(k)} = [\mathbf{v}_1, \dots, \mathbf{v}_k]^T \mathbf{z}_T.
 \label{eq:pca_projection}
\end{equation}
This provides a principled foundation for a curriculum that introduces knowledge of increasing complexity, forming the core of our adaptive distillation strategy.

\section{Method}
\label{sec:method}
\subsection{Overview of \texttt{FAPD}}
\label{sec:overview}
The \texttt{FAPD} framework, illustrated in \autoref{fig:framework_overview}, orchestrates adaptive knowledge transfer through three stages. First, the server performs Hierarchical Knowledge Decomposition (HKD) on the teacher's high-dimensional features $\mathbf{z}_T \in \mathbb{R}^D$ using PCA, generating an orthogonal rotation matrix $R$ that orders dimensions by variance. Second, a Consensus-Driven Curriculum (CDC) controller dynamically adjusts the knowledge complexity $k_t$ for each round $t$. By monitoring the global accuracy history $\mathcal{A}$, the controller evaluates a stability condition $C(t)$ to determine if the network has reached a consensus, advancing $k_t$ when stability is confirmed. Finally, during client-side Progressive Knowledge Distillation (PKD), each client projects its local features $\mathbf{z}_{S,c}$ and the teacher's features into the current $k_t$-dimensional subspace using the projection matrix $P_t$. Clients then optimize a composite objective $\mathcal{L}_c$ that aligns these projected representations, ensuring that knowledge transfer is synchronized with the network's collective learning capacity.

\subsection{Hierarchical Knowledge Decomposition}
To structure the knowledge transfer, we decompose the teacher's $D$-dimensional feature space into an ordered hierarchy. We apply PCA to a representative set of teacher feature embeddings, $\{\mathbf{z}_{T,i} \in \mathbb{R}^D\}_{i=1}^M$, extracted from a calibration dataset, where $M$ denotes the number of calibration samples. The data covariance matrix is:
\begin{equation}
    \Sigma = \frac{1}{M-1} \sum_{i=1}^{M} (\mathbf{z}_{T,i} - \bar{\mathbf{z}}_T)(\mathbf{z}_{T,i} - \bar{\mathbf{z}}_T)^T,
\end{equation}
where $\bar{\mathbf{z}}_T$ is the mean feature vector. Eigendecomposition of $\Sigma$ yields a set of orthogonal principal components:
\begin{equation}
    \Sigma = V \Lambda V^T,
\end{equation}
where $V = [\mathbf{v}_1, \dots, \mathbf{v}_D]$ is the matrix of eigenvectors (principal components) and $\Lambda$ is a diagonal matrix of eigenvalues $\{\lambda_1, \dots, \lambda_D\}$, sorted such that $\lambda_1 \ge \dots \ge \lambda_D$. Consequently, the matrix $R = V^T \in \mathbb{R}^{D \times D}$ serves as a global rotation matrix. At curriculum stage $t$, the projection matrix for dimensionality $k_t$ is constructed by selecting the first $k_t$ rows of $R$:
\begin{equation}
 P_t = R[:k_t, :] \in \mathbb{R}^{k_t \times D}.
 \label{eq:projection_matrix}
\end{equation}
Consequently, this matrix projects any feature vector onto the $k_t$-dimensional subspace that captures the most data variance.

\subsection{Consensus-Driven Curriculum Controller}
The central innovation of \texttt{FAPD} is the server-side curriculum controller that dynamically paces knowledge transfer, where the controller adjusts the feature dimension $k_t$ based on a network-wide stability consensus.

At the end of each round $t$, the server evaluates the global accuracy, $acc_t$, and the network is considered to have reached consensus if the accuracy has plateaued. Formally, let $\epsilon$ be a stability threshold and $N$ be the consensus window size; the stability condition $C(t)$ is met if all recent accuracies are close to the current accuracy:
\begin{equation}
 C(t) = \bigwedge_{\tau=t-N+1}^{t-1} (|acc_t - acc_\tau| < \epsilon).
 \label{eq:stability_condition}
\end{equation}
The curriculum dimension for the next round, $k_{t+1}$, is then updated according to the rule:
\begin{equation}
 k_{t+1} = \begin{cases} \min(k_t + \Delta k, D) & \text{if } C(t) \text{ is true} \\ k_t & \text{otherwise} \end{cases},
 \label{eq:curriculum_update}
\end{equation}
where $k_0$ is the initial dimension and $\Delta k$ is the step size. By implementing this curriculum control, more complex knowledge is introduced only when the network has collectively mastered the current representations.

\subsection{Client-Side Progressive Distillation}
During local training at round $t$, each client $c$ receives the global model parameters $w_t$ and curriculum dimension $k_t$. For each input $\mathbf{x}$, the client computes its feature representation $\mathbf{z}_{S,c} = g_S(\mathbf{x}; w_{S,c})$. To align with the teacher, features are projected into the $k_t$-dimensional subspace using $P_t$ from \autoref{eq:projection_matrix}, yielding $\mathbf{z}_{S,c}^{k_t} = P_t \mathbf{z}_{S,c}$ and $\mathbf{z}_T^{k_t} = P_t \mathbf{z}_T$, where $\mathbf{z}_T$ denotes the teacher's feature representation for the same input.
The local training objective for client $c$ combines three components:
\begin{equation}
    \mathcal{L}_c = \mathcal{L}_{CE} + \lambda_{kd} \mathcal{L}_{KD} + \lambda_{cl} \mathcal{L}_{CL}.
 \label{eq:total_loss}
\end{equation}
Here, the classification loss $\mathcal{L}_{CE}$ is the standard cross-entropy on ground-truth labels, while the knowledge distillation loss $\mathcal{L}_{KD}$ enforces consistency between student and teacher feature distributions in the projected space. First, both feature vectors are L2-normalized; then, we compute the KL-divergence:
\begin{equation}
    \mathcal{L}_{KD} = D_{KL}(\log\text{softmax}(\hat{\mathbf{z}}_{S,c}^{k_t}) \Vert \text{softmax}(\hat{\mathbf{z}}_T^{k_t})),
\end{equation}
where $\hat{\mathbf{z}} = \mathbf{z} / \Vert\mathbf{z}\Vert_2$. The contrastive loss $\mathcal{L}_{CL}$ refines the feature space by aligning image features with text-based semantic embeddings, following the InfoNCE framework adapted from recent diffusion-based federated approaches \cite{wang2025feddifrc}. Let $\mathbf{z}_{\text{text},y}^{k_t}$ be the projected text feature for the correct class $y$ (positive) and $\{\mathbf{z}_{\text{text},j}^{k_t}\}_{j \neq y}$ be for other classes (negatives). The loss is:
\begin{equation}
    \mathcal{L}_{CL} = -\log \frac{\exp(\text{sim}(\mathbf{z}_{S,c}^{k_t}, \mathbf{z}_{\text{text},y}^{k_t})/\tau)}{\sum_{j} \exp(\text{sim}(\mathbf{z}_{S,c}^{k_t}, \mathbf{z}_{\text{text},j}^{k_t})/\tau)},
 \label{eq:infonce_loss}
\end{equation}
where $\tau$ is a temperature hyperparameter and $\text{sim}(\cdot, \cdot)$ is the cosine similarity. Through this progressive, multi-faceted objective, clients learn a rich representation aligned with both visual and semantic features. 
The complete \texttt{FAPD} training procedure is detailed in \autoref{alg:fapd} of the supplementary material.

\section{Experiments}
\label{sec:experiments}

\begin{table}[t]
\caption{Accuracy (\%) of comparison across datasets ($\alpha$=0.5).}
\label{tab:main_accuracy}
\setlength{\tabcolsep}{3.5mm}{
\resizebox{\columnwidth}{!}{
\begin{tabular}{lccc}
\toprule
\rowcolor{gray!8}
\textbf{Method} & CIFAR-10 & CIFAR-100 & Tiny-ImageNet \\
\cmidrule(lr){1-1} \cmidrule(lr){2-4}
FedAvg~\cite{mcmahan2017communicationefficient} & 85.78 & 61.26 & 43.35 \\
FedProx~\cite{li2020federated} & 85.68 & 61.40 & 43.48 \\
MOON~\cite{li2021modelcontrastive} & 86.10 & 61.48 & 43.78 \\
FedNH~\cite{dai2023tackling} & 86.25 & 61.43 & 44.12 \\
FedRCL~\cite{seo2024relaxed} & 86.89 & 62.36 & 44.89 \\
FedCDA~\cite{wang2023fedcda} & 87.11 & 61.95 & 44.56 \\
\midrule
\texttt{FAPD}$_{nadpt}$ & 87.23 & 62.18 & 44.67 \\
\texttt{FAPD}$_{ncont}$ & 87.89 & 63.05 & 45.28 \\
\rowcolor{lightkeycolor}
FAPD (Ours) & \textbf{89.42} & \textbf{63.84} & \textbf{45.35} \\
\bottomrule
\end{tabular}
}}
\vspace{-10px}
\end{table}

\subsection{Experimental Setup}
\label{subsec:setup}
\noindent\textbf{Datasets and Federated Setting.} Experiments are conducted on three benchmark datasets: CIFAR-10/100~\cite{krizhevsky2010convolutional} and Tiny-ImageNet~\cite{le2015tiny}, a 200-class subset of ImageNet with $64 \times 64$ images. The federated environment consists of 10 clients, with 5 randomly selected in each of the 100 communication rounds. To simulate statistical heterogeneity, training data is partitioned among clients using a Dirichlet distribution with concentration parameter $\alpha \in \{0.1, 0.5, 1.0\}$ controlling non-IID intensity, where smaller values induce more severe distribution skew.

\noindent\textbf{Implementation Details.} All experiments are conducted using PyTorch on NVIDIA 4090 GPUs. Client models employ ResNet-10 for CIFAR-10 and CIFAR-100, while MobileNetV2 is used for Tiny-ImageNet to ensure fair comparison. The teacher model provides pre-extracted 512-dimensional features from a Stable Diffusion~\cite{rombach2022highresolution} encoder. Each client trains for $E \in \{5, 10, 20\}$ local epochs per round using SGD optimizer with learning rate 0.01, momentum 0.9, and batch size 64. \texttt{FAPD} is configured with initial curriculum dimension $k_0=8$, step size $\Delta k=5$, stability threshold $\epsilon=0.005$, and consensus window $N=3$ rounds. The distillation objective combines KL-divergence loss and InfoNCE contrastive loss with temperature $\tau=0.04$, weighted by $\lambda_{kd}=0.5$ and $\lambda_{cl}=0.5$.

\noindent\textbf{Baseline Methods.} We compare \texttt{FAPD} against eight baselines: FedAvg~\cite{mcmahan2017communicationefficient}, the standard federated averaging algorithm; FedProx~\cite{li2020federated}, adding proximal terms for heterogeneity handling; MOON~\cite{li2021modelcontrastive}, employing model-level contrastive learning; FedNH~\cite{dai2023tackling}, tackling heterogeneity using class prototypes; FedRCL~\cite{seo2024relaxed}, proposing relaxed contrastive learning; FedCDA~\cite{wang2023fedcda}, aggregating cross-round local models; variant model \texttt{FAPD}$_{nadpt}$, ablation without adaptive mechanism; and variant model \texttt{FAPD}$_{ncont}$, ablation without contrastive learning.

\begin{figure}[t]
\centering
\includegraphics[width=\columnwidth]{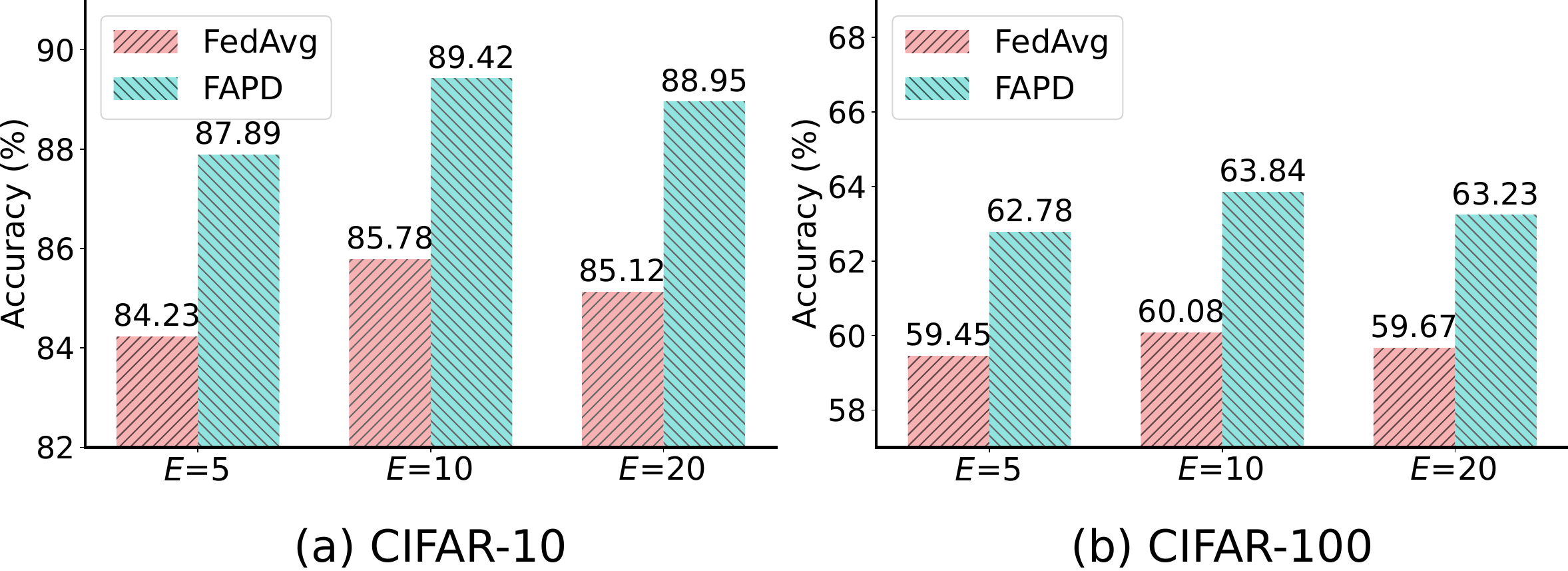}
\caption{Result of local epoch count analysis on CIFAR-10/100.}
\label{fig:local_epochs}
\vspace{-15px}
\end{figure}
\begin{table}[t]
\caption{Ablation study on components and loss functions.}
\label{tab:ablation}
\centering
\setlength{\tabcolsep}{1.5mm}
\resizebox{\columnwidth}{!}{
\begin{tabular}{ccccccccc}
\toprule
\rowcolor{gray!8}
\textbf{ID} & HKD & CDC & PKD & $\mathcal{L}_{CE}$ & $\mathcal{L}_{KD}$ & $\mathcal{L}_{CL}$ & CIFAR-10 & CIFAR-100 \\
\cmidrule(lr){1-1} \cmidrule(lr){2-4} \cmidrule(lr){5-7} \cmidrule(lr){8-9}
1 & \ding{55} & \ding{55} & \ding{55} & \ding{51} & \ding{55} & \ding{55} & 85.78 & 61.26 \\
2 & \ding{55} & \ding{55} & \ding{55} & \ding{51} & \ding{55} & \ding{51} & 86.23 & 61.67 \\
3 & \ding{51} & \ding{55} & \ding{55} & \ding{51} & \ding{51} & \ding{55} & 86.12 & 61.74 \\
4 & \ding{51} & \ding{55} & \ding{55} & \ding{51} & \ding{51} & \ding{51} & 87.23 & 62.18 \\
5 & \ding{51} & \ding{51} & \ding{51} & \ding{51} & \ding{51} & \ding{55} & 87.89 & 63.05 \\
\rowcolor{lightkeycolor}
6 & \ding{51} & \ding{51} & \ding{51} & \ding{51} & \ding{51} & \ding{51} & \textbf{89.42} & \textbf{63.84} \\
\bottomrule
\end{tabular}}
\vspace{-10px}
\end{table}
\begin{figure}[t]
\centering
\includegraphics[width=\columnwidth]{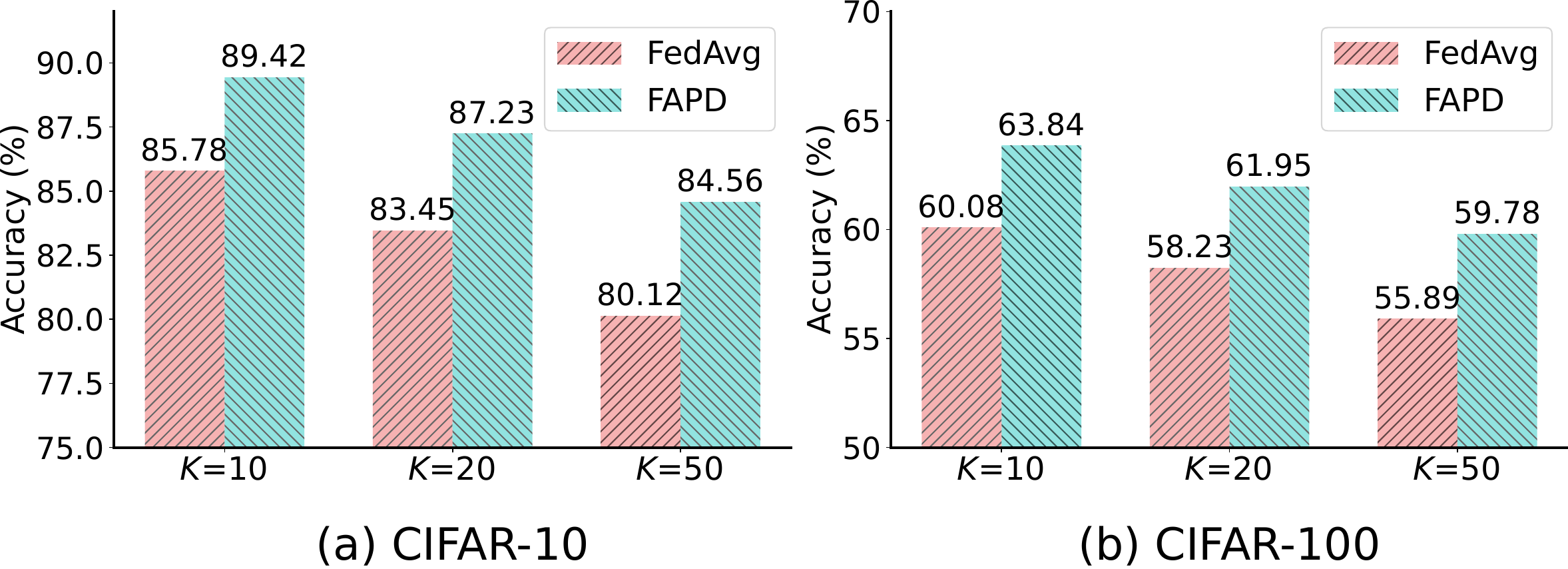}
\caption{Result of client scalability analysis on CIFAR-10/100.}
\label{fig:local_clients}
\vspace{-18px}
\end{figure}

\subsection{Results and Analysis}
\label{subsec:results}
\noindent\textbf{Performance Comparison.} \texttt{FAPD} demonstrates superior performance across all benchmarks, as shown in \autoref{tab:main_accuracy}. On CIFAR-10, it achieves 89.42\% accuracy, outperforming the strongest baseline, FedCDA, by 2.31\% and FedAvg by 3.64\%. This advantage extends to more complex tasks, with \texttt{FAPD} attaining 63.84\% on CIFAR-100 and 45.35\% on Tiny-ImageNet, consistently surpassing contrastive methods like FedRCL and MOON. Ablation results highlight the synergy of our approach: removing the adaptive mechanism (\texttt{FAPD}$_{nadpt}$) or contrastive learning (\texttt{FAPD}$_{ncont}$) leads to significant performance drops of 2.19\% and 1.53\% on CIFAR-10, respectively. These results confirm that \texttt{FAPD}'s adaptive curriculum effectively manages knowledge complexity, yielding substantial gains over static distillation strategies, particularly as dataset difficulty increases.

\noindent\textbf{Scalability Analysis.} We analyze the impact of local epochs ($E$) and client scalability ($K$) on convergence and accuracy, as illustrated in \autoref{fig:local_epochs} and \autoref{fig:local_clients}. For local epochs, \texttt{FAPD} achieves optimal performance at $E=10$ on both CIFAR-10 (89.42\%) and CIFAR-100 (63.84\%). While increasing $E$ to 20 slightly degrades accuracy (e.g., 88.95\% on CIFAR-10) due to potential client drift, \texttt{FAPD} consistently outperforms FedAvg across all $E \in \{5, 10, 20\}$, demonstrating robustness to local training variations. Regarding scalability, increasing the total client count $K$ from 10 to 50 introduces greater heterogeneity and data sparsity. Consequently, accuracy naturally declines, yet \texttt{FAPD} maintains a significant lead. On CIFAR-10, \texttt{FAPD} achieves 84.56\% with $K=50$, surpassing FedAvg's 80.12\% by 4.44\%. Similarly, on CIFAR-100, \texttt{FAPD} retains 59.78\% accuracy at $K=50$ versus FedAvg's 55.89\%, validating the framework's effectiveness in larger-scale federated networks.

\noindent\textbf{Ablation Study.} 
\autoref{tab:ablation} systematically isolates \texttt{FAPD}'s component contributions. Model 1 (FedAvg/baseline) with only $\mathcal{L}_{CE}$ achieves 85.78\% on CIFAR-10. Model 2 adds $\mathcal{L}_{CL}$, improving to 86.23\%, demonstrating semantic alignment benefits. Model 3 introduces HKD with $\mathcal{L}_{KD}$, reaching 86.12\%, while Model 4 (\texttt{FAPD}$_{nadpt}$) combines HKD with both distillation losses, achieving 87.23\%. Model 5 (\texttt{FAPD}$_{ncont}$) integrates all components (HKD, CDC, PKD) but excludes contrastive learning, reaching 87.89\% and highlighting consensus-driven adaptive pacing effectiveness. Model 6 (complete \texttt{FAPD}) achieves peak performance of 89.42\% on CIFAR-10 and 63.84\% on CIFAR-100. Comparing Models 5 and 6 reveals $\mathcal{L}_{CL}$ contributes 1.53\% gain, while Models 4 and 5 show adaptive curriculum control (CDC+PKD) adds 0.66\%. The synergistic combination enables \texttt{FAPD} to effectively balance knowledge complexity with client learning capacity across heterogeneous federated environments.

\begin{figure}[t]
\centering
\includegraphics[width=\columnwidth]{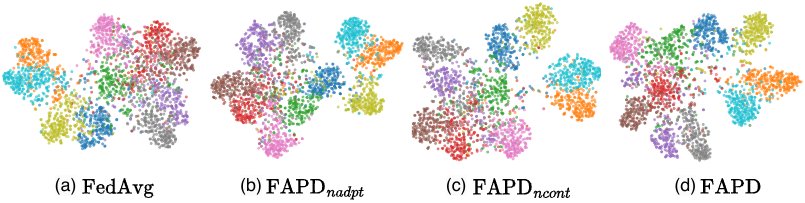}
\caption{t-SNE visualization comparing feature distributions.}
\label{fig:tsne}
\vspace{-15px}
\end{figure}

\noindent\textbf{Visualization Analysis.} 
\autoref{fig:tsne} visualizes the t-SNE embeddings of CIFAR-10 test samples to assess representation quality. FedAvg results in severely entangled clusters with ambiguous boundaries, particularly between similar classes like automobiles and trucks. In contrast, \texttt{FAPD} produces compact, well-separated clusters with distinct inter-class margins. While the non-adaptive variant (\texttt{FAPD}$_{nadpt}$) shows only slight improvement, the full \texttt{FAPD} framework effectively organizes the feature space, aligning closely with semantic hierarchies. Quantitatively, \texttt{FAPD} achieves 3.64 percentage points higher accuracy than FedAvg on CIFAR-10 (89.42\% vs. 85.78\%), representing a 4.24\% relative improvement, demonstrating that the synergistic combination of hierarchical decomposition and consensus-driven adaptation enables the learning of highly discriminative representations that capture both coarse and fine-grained semantic distinctions.

\noindent\textbf{Robustness Analysis.}
We evaluate \texttt{FAPD}'s resilience to statistical heterogeneity using Dirichlet partitions ($\alpha \in \{0.1, 0.2, 0.5, 1.0\}$), as shown in \autoref{tab:heterogeneity}. Under extreme non-IID conditions ($\alpha=0.1$), \texttt{FAPD} maintains 85.87\% accuracy on CIFAR-10, outperforming FedAvg (81.35\%) by 4.52\%. Notably, as heterogeneity increases ($\alpha$ from 1.0 to 0.1), FedAvg suffers a sharp 5.77\% drop, whereas \texttt{FAPD} degrades by only 4.31\%, demonstrating superior stability. This robustness extends to CIFAR-100, where \texttt{FAPD} achieves 60.12\% at $\alpha=0.1$, surpassing FedAvg's 56.12\% by 4.00\%. The consistent performance advantage of the full \texttt{FAPD} over variants (\texttt{FAPD}$_{nadpt}$ and \texttt{FAPD}$_{ncont}$) across all $\alpha$ settings confirms that the consensus-driven adaptive curriculum effectively mitigates client drift caused by severe distribution skew.

\begin{table}[t]
\caption{Accuracy (\%) under varying $\alpha$ level.}
\label{tab:heterogeneity}
\setlength{\tabcolsep}{2.5mm}
\resizebox{\columnwidth}{!}{
\begin{tabular}{cccccc}
\toprule
\rowcolor{gray!8}
\textbf{Dataset} & \textbf{Method} & $\alpha=0.1$ & $\alpha=0.2$ & $\alpha=0.5$ & $\alpha=1.0$ \\
\cmidrule(lr){1-1} \cmidrule(lr){2-2} \cmidrule(lr){3-6}
\multirow{4}{*}{\rotatebox{90}{CIFAR-10}}
& FedAvg~\cite{mcmahan2017communicationefficient} & 81.35 & 84.12 & 85.78 & 87.12 \\
& \texttt{FAPD}$_{nadpt}$ & 83.12 & 85.78 & 87.23 & 88.45 \\
& \texttt{FAPD}$_{ncont}$ & 84.56 & 86.89 & 87.89 & 89.12 \\
& \cellcolor{lightkeycolor}FAPD (Ours) & \cellcolor{lightkeycolor}\textbf{85.87} & \cellcolor{lightkeycolor}\textbf{88.34} & \cellcolor{lightkeycolor}\textbf{89.42} & \cellcolor{lightkeycolor}\textbf{90.18} \\
\midrule
\multirow{4}{*}{\rotatebox{90}{CIFAR-100}}
& FedAvg~\cite{mcmahan2017communicationefficient} & 56.12 & 58.45 & 61.26 & 62.89 \\
& \texttt{FAPD}$_{nadpt}$ & 57.45 & 59.78 & 62.18 & 63.56 \\
& \texttt{FAPD}$_{ncont}$ & 58.89 & 61.23 & 63.05 & 64.45 \\
& \cellcolor{lightkeycolor}FAPD (Ours) & \cellcolor{lightkeycolor}\textbf{60.12} & \cellcolor{lightkeycolor}\textbf{62.45} & \cellcolor{lightkeycolor}\textbf{63.84} & \cellcolor{lightkeycolor}\textbf{65.12} \\
\bottomrule
\end{tabular}
}
\end{table}

\noindent\textbf{Discussion}.
While \texttt{FAPD} demonstrates significant efficacy, several limitations warrant discussion. First, the hierarchical decomposition relies on PCA derived from a calibration dataset. If this data does not adequately represent the global distribution, the resulting knowledge hierarchy may be suboptimal. Second, the current consensus mechanism applies a unified curriculum across the network. In scenarios with extreme resource heterogeneity, a personalized curriculum that adapts to individual client capacities could potentially offer further gains. Finally, our framework is currently optimized for image classification tasks using CNNs. Extending the hierarchical decomposition principle to other multimedia modalities, such as video sequences or audio spectrograms, requires domain-specific adaptations of the variance-based ordering.

\section{Conclusions}
\label{sec:conclusion}
We present \texttt{FAPD}, a consensus-driven framework that dynamically orchestrates knowledge transfer complexity in collaborative knowledge distillation. By monitoring network-wide learning stability and progressively expanding feature dimensionality through PCA-based hierarchical decomposition, \texttt{FAPD} adaptively advances curriculum dimensionality when collective consensus emerges, ensuring knowledge complexity matches heterogeneous client capacity. Extensive experiments on three datasets demonstrate \texttt{FAPD} achieves 3.64\% accuracy improvement over FedAvg with 2$\times$ faster convergence, and exhibits remarkable resilience to data heterogeneity, maintaining high performance even under extreme non-IID settings. Future work will explore client-specific adaptive curricula and multi-modal extensions to video analytics tasks.
\bibliographystyle{IEEEtran}
\begin{spacing}{0.95}
\bibliography{refs}
\end{spacing}

\newpage
\end{document}